# Decisions with Limited Observations over a Finite Product Space: The Klir Effect

Michael Pittarelli
Computer Science Department, SUNY Institute of Technology
P.O. Box 3050, Utica, NY 13504-3050

**Abstract**

Probability estimation by maximum entropy reconstruction of an initial relative frequency estimate from its projection onto a hypergraph model of the approximate conditional independence relations exhibited by it is investigated. The results of this study suggest that use of this estimation technique may improve the quality of decisions that must be made on the basis of limited observations over a decomposable finite product space.

## I. Introduction

Since the mid-1970s, algorithms have been available for identifying hypergraph models of the conditional independence relations holding among a set of finite random variables over which a probability distribution (usually estimated as an observed relative frequency distribution) is given. The independence relations implied by the structure of a model hold iff the distribution coincides with the maximum entropy extension (reconstruction) of its projections onto (marginals over) the components of the model; and they may be "read from" the hypergraph model as they may be from Markov networks [1] or graphical models [2] (which hypergraph models include as a special case).

When a distribution approximately coincides with its reconstruction from a model, at least three interpretations are possible:

1. The specified independence relations among the variables hold approximately.

2. The relations hold absolutely, but the given distribution is merely an approximation to the true distribution.

3. Both the distribution and the model are approximations.

The model search procedures referred to above have been applied successfully in a number of exploratory studies in which dependencies were not known a priori [3].

In the course of testing the accuracy of the procedures using distributions over sets of variables with known dependencies, it was discovered that, on the average, the relative frequency estimate from observations generated in accordance with the true distribution diverged more from the true distribution than did the distribution reconstructed from the projection of the estimate onto the model identified by the search procedure as best for the estimate [4]. Later experiments using randomly generated distributions with no known dependencies established that the average distance of the reconstruction from the true distribution was smaller when the number of observations was relatively small (the actual number depending on the number of variables involved and the number of values taken by them) [5].

The purpose of this paper is to demonstrate the appropriateness of the probability estimation technique suggested by these experiments for decision problems in which probabilities (over a finite, decomposable product space) must be estimated quickly from observed relative frequencies. Results will be presented for sets of variables exhibiting both exact and approximate dependency relations. It will also be shown that this estimation method belongs to a recently identified category of statistical smoothing techniques.

## II. Definitions and Notation

A probability distribution over a finite product space is specified by the table below:

| $v_1$ | $v_2$ | $v_3$ | p(.) |
|---|---|---|---|
| 0 | 0 | 0 | 7/32 |
| 0 | 0 | 1 | 1/32 |
| 0 | 1 | 0 | 7/64 |
| 0 | 1 | 1 | 1/64 |
| 1 | 0 | 0 | 21/64 |
| 1 | 0 | 1 | 3/64 |
| 1 | 1 | 0 | 7/32 |
| 1 | 1 | 1 | 1/32 |

$V=\{v_1,v_2,v_3\}$; $dom(V)=dom(v_1)\times dom(v_2)\times dom(v_3)=\{0,1\}^3$.

The *projection* of p onto $\{v_1,v_2\}$, denoted $\pi_{\{v_1,v_2\}}(p)$, is

| $v_1$ | $v_2$ | $\pi_{\{v_1,v_2\}}(p)(.)$ | |
|---|---|---|---|
| 0 | 0 | 1/4 | (= 7/32 + 1/32) |
| 0 | 1 | 1/8 | (= 7/64 + 1/64) |
| 1 | 0 | 3/8 | (= 21/64 + 3/64) |
| 1 | 1 | 1/4 | (= 7/32 + 1/32) |

A (*hypergraph*) *model* of a set V of variables is any collection $X=\{V_1, \ldots, V_m\}$ such that

1. $\cup_{i=1}^{m} V_i = V$;
2. $i\neq j$ implies $V_i \not\subseteq V_j$.

(It is sometimes useful to consider a wider class of models satisfying condition 2 and
1′. $\cup_{i=1}^{m} V_i \subseteq V$.)

The projection of p onto a model $X=\{V_1, \ldots, V_m\}$ is the set of distributions

$$\pi_X(p)=\{\pi_{V_1}(p), \ldots, \pi_{V_m}(p)\}.$$

The *extension* of a set $\pi_X(p)$ is the set

$$E(\pi_X(p))=\{p'|\pi_X(p')=\pi_X(p)\}.$$

The *maximum entropy extension* of $\pi_X(p)$ is the distribution $J(\pi_X(p))\in E(\pi_X(p))$ such that

$$H(J(\pi_X(p)))=\max_{p'\in E(\pi_X(p))} H(p'),$$

where

$$H(p)=-\sum_{s\in dom(V)} p(s)\log(p(s)).$$

(This distribution is unique and easily calculated.) If $p=J(\pi_X(p))$, then p is *reconstructable from* X.

## III. Model Search

A partial ordering, *refinement*, is defined on the set M of all models over V as $X\leq Y$ iff for every $V_x\in X$ there exists a $V_y\in Y$ such that $V_x\subseteq V_y$ [6]. The pair $(M,\leq)$ is a lattice with greatest element $\{V\}$ and least element $\{\{v\}|v\in V\}$ [7]. (For the extended set of models, the least element is $\{\emptyset\}$.) The figure below is a Hasse diagram of the lattice of models over $\{v_1,v_2,v_3\}$:

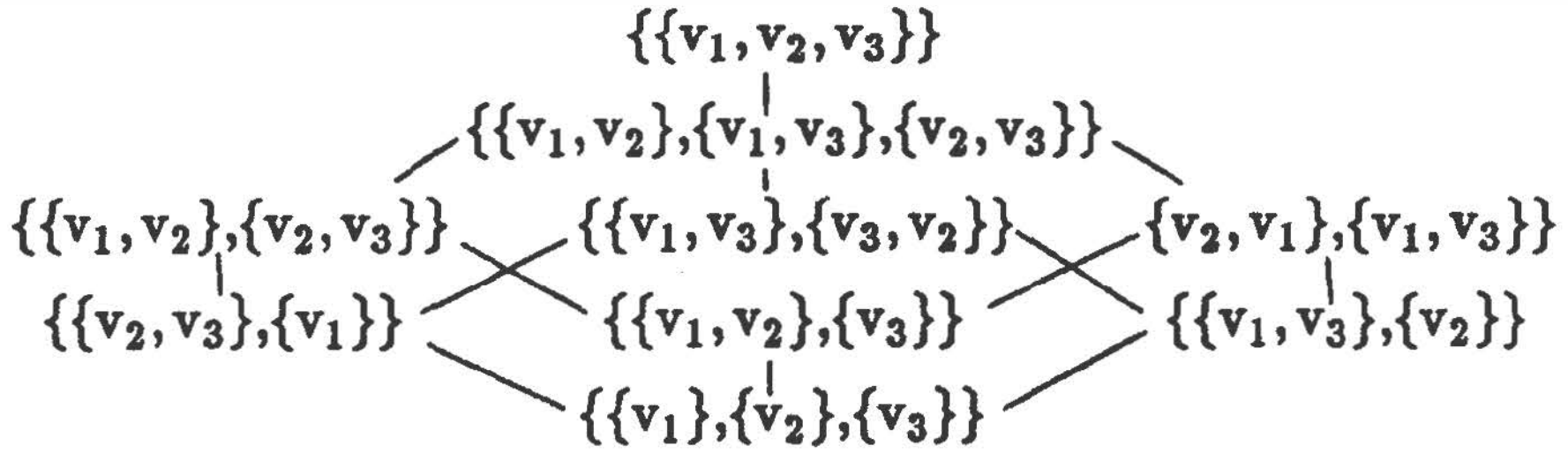

The models may be more perspicuously represented as block diagrams [6]. For example, $\{\{v_1,v_2\},\{v_2,v_3\}\}$ is represented as

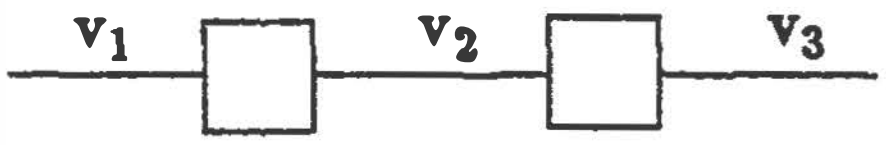

A distribution is reconstructable from this model iff $v_1$ and $v_3$ are conditionally independent, given $v_2$ [8].

Two (opposing) criteria are used to assess the relative quality of models of a distribution p:

1. All else being equal, X is preferred to Y if its distance in the lattice from $\{\{v\}|v\in V\}$ is less.
2. All else being equal, X is preferred to Y if

$$d(p,J(\pi_X(p))) \le d(p,J(\pi_Y(p))),$$

where d(p,q) is the directed divergence (cross-entropy, relative entropy) from p to q:

$$d(p,q) = \sum_s p(s) \log(p(s)/q(s)).$$

A simplified version of a search procedure that has worked well in experiments involving randomly generated distributions with known dependencies is roughly as follows:

$X_0$:={V}
**repeat**
  $X_{i+1}$:= immediate refinement Y of $X_i$
    such that $d(p,J(\pi_Y(p)))$ is minimum
**until** stopping criterion is met

The stopping criterion is usually defined in terms of the differences

$$d(p,J(\pi_{X_{i+1}}(p))) - d(p,J(\pi_{X_i}(p))).$$

## IV. The Klir Effect

Let $\hat{p}_n$ denote a relative frequency estimate of a distribution p in accordance with which n observations have been generated randomly. What was discovered by George Klir and his co-workers in the experiments [4] referred to above is that on the average, if $p=\pi_X(p)$, then

$$h(p,J(\pi_X(\hat{p}_n))) < h(p,\hat{p}_n),$$

where h is Hamming distance (sum of absolute deviations; directed divergence is not defined for all pairs of distributions). For random p without known dependencies, where X is identified as the best model of the (approximate) dependencies expressed by $\hat{p}_n$, the inequality holds for relatively small values of n [5]. Thus, in any case, for small n, $J(\pi_X(\hat{p}_n))$ is usually a more accurate estimate of p than is $\hat{p}_n$.

To illustrate, the table below gives average Hamming distances $h(p,\hat{p}_n)$ and $h(p, J(\pi_X(\hat{p}_n)))$ for 1000 distributions p randomly generated so as to be reconstructable from the model (Markov network) $X=\{\{v_1,v_2\},\{v_3\}\}$.

| n | $h(p,\hat{p}_n)$ | $h(p,J(\pi_X(\hat{p}_n)))$ |
|---|---|---|
| 5 | 0.9476 | 0.7307 |
| 10 | 0.6492 | 0.5030 |
| 40 | 0.3220 | 0.2489 |
| 500 | 0.0930 | 0.0712 |
| 5000 | 0.0284 | 0.0219 |
| 50000 | 0.0090 | 0.0068 |

On the average,

(1) for each n,

$$h(p,\hat{p}_n) > h(p, J(\pi_X(\hat{p}_n)));$$

(2) for $n > n'$,

(a) $$h(p,\hat{p}_n) < h(p,\hat{p}_{n'}) \text{ and}$$

(b) $$h(p, J(\pi_X(\hat{p}_n))) < h(p, J(\pi_X(\hat{p}_{n'}))).$$

Observation (2a) requires no explanation; (2b) follows from (2a) and the reconstructability of p from X. Observation (1) is explained as follows. Suppose that p is reconstructable from a model $X=\{V_1, \ldots, V_m\}$. Then

$$p=J(\{\pi_{V_1}(p), \ldots, \pi_{V_m}(p)\}).$$

For any value of n, $\hat{p}_n$ is an approximation to p and $\pi_{V_i}(\hat{p}_n)$ is an approximation to $\pi_{V_i}(p)$. However, $\pi_{V_i}(\hat{p}_n)$ is a *better* approximation than is $\hat{p}_n$, because

$$n/|\mathrm{dom}(V_i)| \;>\; n/|\mathrm{dom}(V_1 \cup \cdots \cup V_m)|.$$

Therefore, $J(\pi_X(\hat{p}_n))$ tends to be a better estimate of p than $\hat{p}_n$ is [9].

I shall refer to this phenomenon as *the Klir effect.* Estimation of p as $J(\pi_X(\hat{p}_n))$ I shall refer to as *Klir estimation.*

(Apparently unknown to Klir and his co-workers was the experiment described in Bishop's 1967 Ph.D. dissertation [10]. Although her main concern is to eliminate sampling zeros and minimize the variance of cell estimates for cross-classified frequency data, Bishop reports the results of an experiment identical in form to those discussed above. For a variable $v_1$ with 3 categories and dichotomous variables $v_2$ and $v_3$, and model $X=\{\{v_1,v_2\},\{v_2,v_3\},\{v_1,v_3\}\}$, a frequency distribution $\hat{p}_n$, with n=1000, was generated in accordance with each of 100 randomly generated p such that $p=J(\pi_X(p))$. Bishop reports that the divergence between the true and reconstructed frequency distributions was smaller in 99 of the 100 cases than that between the true and originally generated frequencies. Klir el al. [4,5], on the other hand, report results based on over 30,000 distinct experiments involving models and variables of many varied types.)

It is not always the case that exact conditional independence relations will hold for a set of variables relevant to a given inference or decision problem. Therefore, the underlying probability distribution will not always be reconstructable from some (non-trivial) model. However, it follows from arguments of E.T. Jaynes and Herbert Simon that in practice one can expect an unknown distribution over a set of variables under empirical study to be nearly reconstructable.

Simon's arguments for the near-decomposability of natural and artificial systems are well known [11,12]. In the present context, near-decomposability translates into near-independence of certain subsets $V_1, \ldots, V_m$ of V and therefore near-reconstructability of p from $X = \{V_1, \ldots, V_m\}$.

It is a consequence of E.T. Jaynes' *concentration theorem* [13,14] that the underlying probability distribution generating a set of observed marginal relative frequency distributions $\pi_X(\hat{p}_n)$ is likely to be very close to $J(\pi_X(\hat{p}_n))$, which itself rarely coincides with $\hat{p}_n$ [8]. (See [15] for criticism.)

Relative to a set $P_V$ of probability distributions over a set of finite variables V, let

$$O(X) = \{p \in P_V \mid X \text{ is identified as the best model of } p\}.$$

(Recall that selection of an optimal model involves some tradeoff between information retention and model simplicity.) If the model search procedure breaks ties deterministically, then

$$\{O(X) \mid X \text{ is a model of } V\}$$

is a partition of $P_V$.

For an assumed nearly-reconstructable p for which the dependency structure is unknown a priori, Klir estimation involves three steps:

1. Calculate $\hat{p}_n$.
2. Identify X such that $\hat{p}_n \in O(X)$.
3. Estimate p as $J(\pi_X(\hat{p}_n))$.

The success of this procedure seems to be a consequence of the fact that a large $h(p, \hat{p}_n)$ does not preclude p and $\hat{p}_n$ being in the same equivalence class O(X).

To illustrate, the table below gives percentages with which the Klir estimate is closer to the true distribution p than is $\hat{p}_n$ for three sets of 1000 nearly-reconstructable p, produced by perturbing values of distributions $p_0$ perfectly reconstructable from a randomly selected model of $V = \{v_1, v_2, v_3\}$, $|dom(v_i)| = 2$, as $p(x) = p_0(x) \pm \epsilon$:

| n | $\epsilon = 0.015$ | $\epsilon = 0.05$ | $\epsilon = 0.1$ |
|---|---|---|---|
| 5 | 98% | 94% | 88% |
| 10 | 90% | 86% | 79% |
| 40 | 84% | 72% | 54% |
| 500 | 72% | 31% | 24% |
| 5000 | 46% | 16% | 18% |

Interestingly, the crossover point at which one does better with $\hat{p}_n$ than with $J(\pi_X(\hat{p}_n))$ is roughly $n = 40$; and for these variables, $40 = 5 \times |dom(V)|$, a commonly recommended minimum number of observations for reliable application of many discrete multivariate statistical procedures.

This type of estimation is a member of a (wide) class of techniques recently identified by Titterington [16]. Let $p_c$ denote the centroid of $P_V$ (i.e., the uniform distribution over dom(V)). A *minimum penalized roughness* estimate is a distribution $\tilde{p} \in P_V$ such that

$$\Delta_1(\tilde{p}, p_c) + k\Delta_2(\tilde{p}, \hat{p}_n)$$

is minimized, where $k>0$ and the $\Delta_i: P_V \times P_V \rightarrow \mathbb{R}$ needn't be metric distances but are usually nondegenerate. One candidate for $\Delta_1$ suggested by Titterington is directed divergence (Section III). It is easy to show [8] that the unique $p^* \in E(\pi_X(\hat{p}_n))$ such that

$$d(p^*, p_c) = \min_{p \in E(\pi_X(\hat{p}_n))} d(p, p_c)$$

is also the unique p* such that

$$H(p^*) = \max_{p \in E(\pi_X(\hat{p}_n))} H(p),$$

viz., $J(\pi_X(\hat{p}_n))$. Let

$$\Delta_2(p, \hat{p}_n) = \begin{cases} 0, & \text{if } p \in E(\pi_X(\hat{p}_n)), \text{ where } \hat{p}_n \in O(X) \\ \infty, & \text{otherwise} \end{cases}$$

Then estimating p as $\tilde{p} = J(\pi_X(\hat{p}_n))$ is equivalent to minimizing

$$\Delta_1(\tilde{p}, p_c) + k\Delta_2(\tilde{p}, \hat{p}_n)$$

with $\Delta_1 = d$, $\Delta_2$ as defined above, and k=1.

The more refined X is the greater the degree of "smoothing" of the estimate, since $X \leq Y$ implies

$$E(\pi_Y(\hat{p}_n)) \subseteq E(\pi_X(\hat{p}_n)),$$

which in turn implies

$$d(J(\pi_X(\hat{p}_n)), p_c) \leq d(J(\pi_Y(\hat{p}_n)), p_c).$$

At one extreme, $X = \{\emptyset\}$ gives $\tilde{p} = p_c$. At the other, $X = \{V\}$ gives $\tilde{p} = \hat{p}_n$ [17]. (See also Chapter 12 and Section 14.7 of Bishop, Fienberg, and Holland [18].)

### V. Decisions and Klir Estimation

It may be argued that the ultimate purpose of estimating probabilities is to base decisions on them. (See Kyburg [19].) What I have been referring to as the Klir effect would then be of practical significance only if it could be used to improve decision making. Experiments involving randomly generated decision problems and nearly-reconstructable distributions suggest that this is the case.

Consider a decision problem in which the states (conditions) are the 8 possible locations of a moving object within a discrete 3-dimensional grid, dimension i associated with $v_i$, where $dom(v_i) = \{0, 1\}$; and in which there are 8 actions $a_w$, where $a_w$ is the act of reaching into cell w of the grid and grabbing. Imagine that, although the object is completely oblivious to attempts to grab it, it moves so quickly and erratically that reaching into the cell that it is currently observed to occupy is a hopeless strategy. Instead, the decision is based on an estimate of the probability of occupation of each cell (which in turn is based on an observed frequency distribution of occupation as recorded by some monitoring device) and the utility matrix:

| | $s_{v_1v_2v_3=000}$ | $s_{001}$ | $s_{010}$ | $s_{011}$ | $s_{100}$ | $s_{101}$ | $s_{110}$ | $s_{111}$ |
|---|---|---|---|---|---|---|---|---|
| $a_{000}$ | 100 | 0 | 0 | 0 | 0 | 0 | 0 | -50 |
| $a_{001}$ | 0 | 10 | 0 | 0 | 0 | 0 | -10 | 0 |
| $a_{010}$ | 0 | 0 | 10 | 0 | 0 | -5 | 0 | 0 |
| $a_{011}$ | 0 | 0 | 0 | 10 | -5 | 0 | 0 | 0 |
| $a_{100}$ | 0 | 0 | 0 | -5 | 10 | 0 | 0 | 0 |
| $a_{101}$ | 0 | 0 | -5 | 0 | 0 | 10 | 0 | 0 |
| $a_{110}$ | 0 | -10 | 0 | 0 | 0 | 0 | 10 | 0 |
| $a_{111}$ | -30 | 0 | 0 | 0 | 0 | 0 | 0 | 100 |

Suppose that this probability distribution is unknown, but that it is known to be reconstructable from $\{\{v_1\},\{v_2\},\{v_3\}\}$.

For a decision problem with actions $A=\{a_1,\ldots,a_m\}$ and states $S=\{s_1,\ldots,s_n\}=\text{dom}(V)$, let $e_p(a_i)$ denote the expected utility of $a_i$ relative to p. Let

$$D(a_i)=\{p\in P_V \mid e_p(a_i)\geq e_p(a_j); i,j\in\{1,\ldots,m\}\}$$

denote the set of distributions relative to which act $a_i$ maximizes expected utility. Each $D(a_i)$ is a convex polyhedron and the collection $\{D(a_i)\mid a_i\in A\}$ is a cover of $P_V$ (sets $D(a_i)$ and $D(a_j)$ may have a common face) [20].

Relative to a given decision problem, estimate $p^i$ of p (the actual distribution over S) is superior to estimate $p^j$, $p^i > p^j$, iff $p^i$ and p are both elements of some D(a) and for no a∈A are $p^j$ and p both elements of D(a). Thus, it is possible that $p^i > p^j$ even though $h(p^j,p) < h(p^i,p)$.

For the example, 4000 distributions p over $\{v_1,v_2,v_3\}$ reconstructable from $X=\{\{v_1\},\{v_2\},\{v_3\}\}$ were generated and the frequencies with which $\hat{p}_n, p\in D(a)$ and $J(\pi_X(\hat{p}_n)), p\in D(a)$ were determined for various numbers of observations n:

| n | $\hat{p}_n, p\in D(a)$ | $J(\pi_X(\hat{p}_n)), p\in D(a)$ |
|---|---|---|
| 5 | 494 | 617 |
| 10 | 673 | 708 |
| 40 | 794 | 814 |
| 500 | 939 | 951 |
| 5000 | 987 | 991 |

Results for 1000 randomly generated distributions (not necessarily reconstructable) and 16,000 decision matrices are tabulated below. For each of the 8 possible models of three (dichotomous) variables, 125 distributions $p_0$ perfectly reconstructable from that model were generated and perturbed to form a distribution p as $p(x)=p_0(x)\pm\epsilon$, for $\epsilon=0, 0.025, 0.05, 0.1$. Then 4 matrices with 10 acts each were generated for every p. The entries give the differences

$$\%\text{ of cases in which } p, J(\pi_X(\hat{p}_n))\in D(a) - \%\text{ of cases in which } p, \hat{p}_n\in D(a')$$

as a function of n and $\epsilon$, for the model X for which $\hat{p}_n\in O(X)$:

| n | $\epsilon$=0 | $\epsilon$=0.015 | $\epsilon$=0.05 | $\epsilon$=0.1 |
|---|---|---|---|---|
| 5 | 6.5% | 6.2% | 6.4% | 5.3% |
| 10 | 5.6% | 5.9% | 5.4% | 4.8% |
| 40 | 4.6% | 3.9% | 3.2% | 0.8% |
| 500 | 0.8% | 0.3% | −1.4% | −2.3% |
| 5000 | 0.7% | −0.5% | −3.1% | −4.9% |

The more nearly reconstructable from X is p and the smaller n is, the likelier it is that some action maximizes expected utility relative to both $J(\pi_X(\hat{p}_n))$ and p but not relative to $\hat{p}_n$.

## VI. Conclusions

The results of this study suggest that, for inference and decision problems in which conditional independence relations among a set of finite random variables are known to hold, Klir estimation of a joint probability distribution from observed frequencies is unqualifiedly superior to the usual relative frequency estimate. Its superiority is most pronounced when the number of observations is relatively low. Thus, it would seem that the Klir estimation technique would be very useful in situations where a decision must be made quickly, on the basis of limited observations.

When independence relations are not known a priori, but under the assumption that most sets of variables identified as relevant to some actual problem will exhibit such relations at least approximately, Klir estimation using the model identified as best for the relative frequency estimate seems advantageous for small numbers of observations.